# MICROTASK CROWDSOURCING FOR DISEASE MENTION ANNOTATION IN PUBMED ABSTRACTS

BENJAMIN M GOOD

*Molecular and Experimental Medicine, The Scripps Research Institute, 10550 N. Torrey Pines Rd., La Jolla, CA, 92037, USA*
*Email: bgood@scripps.edu*

MAX NANIS

*Molecular and Experimental Medicine, The Scripps Research Institute, 10550 N. Torrey Pines Rd., La Jolla, CA, 92037, USA*
*Email: max@maxnanis.com*

ANDREW I SU

*Molecular and Experimental Medicine, The Scripps Research Institute, 10550 N. Torrey Pines Rd., La Jolla, CA, 92037, USA*
*Email: asu@scripps.edu*

Identifying concepts and relationships in biomedical text enables knowledge to be applied in computational analyses. Many biological natural language process (BioNLP) projects attempt to address this challenge, but the state of the art in BioNLP still leaves much room for improvement. Progress in BioNLP research depends on large, annotated corpora for evaluating information extraction systems and training machine learning models. Traditionally, such corpora are created by small numbers of expert annotators often working over extended periods of time. Recent studies have shown that workers on microtask crowdsourcing platforms such as Amazon's Mechanical Turk (AMT) can, in aggregate, generate high-quality annotations of biomedical text. Here, we investigated the use of the AMT in capturing disease mentions in PubMed abstracts. We used the NCBI Disease corpus as a gold standard for refining and benchmarking our crowdsourcing protocol. After several iterations, we arrived at a protocol that reproduced the annotations of the 593 documents in the 'training set' of this gold standard with an overall F measure of 0.872 (precision 0.862, recall 0.883). The output can also be tuned to optimize for precision (max = 0.984 when recall = 0.269) or recall (max = 0.980 when precision = 0.436). Each document was examined by 15 workers, and their annotations were merged based on a simple voting method. In total 145 workers combined to complete all 593 documents in the span of 1 week at a cost of $.06 per abstract per worker. The quality of the annotations, as judged with the F measure, increases with the number of workers assigned to each task such that the system can be tuned to balance cost against quality. These results demonstrate that microtask crowdsourcing can be a valuable tool for generating well-annotated corpora in BioNLP. Data produced for this analysis are available at http://figshare.com/articles/Disease_Mention_Annotation_with_Mechanical_Turk/1126402.

# 1. Background

A large proportion of all biomedical knowledge is represented in text. There are currently over 23 million articles indexed in PubMed, and over one million new articles are added every year. Natural language processing (NLP) approaches attempt to extract this knowledge in the form of structured concepts and relationships such that it can be used for a variety of computational tasks. Just a few of many examples include identifying functional genetic variants [1], identifying biomarkers and phenotypes related to disease [2], and drug repositioning [3].

Research in NLP is largely organized around shared tasks [4]. Periodically, the community settles on a particular challenge (e.g., identifying genes in abstracts [5]), develops manually annotated corpora that reflect the objective of the challenge, and then organizes competitions meant to identify the best computational methods. These shared corpora make it possible for researchers to refine their predictive models (in particular to train models based on supervised learning) and to evaluate the performance of all approaches. These gold standard annotated corpora are generally produced by small teams of well-trained annotators. While this methodology has been fruitful, the costs inherent to this approach impose limits on the numbers of different corpora as well as the size of individual corpora that can be produced.

Microtask crowdsourcing platforms, such as Amazon's Mechanical Turk (AMT), facilitate transactions between a 'requester' and hundreds of thousands if not millions of 'workers'. These markets make it possible to harness vast amounts human labor in parallel. Typically a requestor sends a long list of small, discrete "Human Intelligence Tasks" (HITs) to the AMT platform which then distributes the HITs to workers. Workers who choose to work on a given task are paid for each HIT they complete at a rate set by the requestor.

Since their inception, microtask markets have attracted the attention of the NLP community because of the well-known costs of creating annotated corpora [6]. These markets are seen as a way of reducing these costs and dramatically extending the potential size of the datasets needed for training and evaluation [7]. This approach has been particularly useful for tasks that are easy for humans and clearly involve no domain knowledge. For example, a large amount of NLP research is devoted to analyzing text for emotional polarity (e.g. happy or sad).

Adoption of these techniques within the biomedical domain has been slower, probably because of the increased complexity of the texts that need to be processed and the concepts that need to be annotated. That being said, a growing number of BioNLP research groups are exploring the use of microtask crowdsourcing. Zhai et al (2013) demonstrated that a gold standard corpus for clinical natural language processing could be assembled through crowdsourcing [8]. They presented workers on the Crowdflower microtask platform (http://crowdflower.com) with text from clinical trial announcements and asked them to 1) highlight medication names and types, 2) correct other workers highlighting, and 3) link medications with their attributes (e.g. dose) for 3 cents a task. By aggregating the contributions of 5 workers per task, they obtained good results including an F measure of 0.87 on the medication name highlighting task and 0.96 for linking medications with their attributes. In related work, Burger et al reported 85% accuracy for AMT workers (in aggregate) on the task of validating predicted gene-mutation relations in MEDLINE abstracts [9].

Here we extend these efforts by implementing and testing a disease mention annotation task on the AMT platform.

## 2. Task: disease mention annotation

A fundamental step in nearly all NLP tasks is the identification of occurrences of concepts such as diseases, genes, or drugs. Our goal for this work was thus to develop and benchmark a crowdsourcing protocol for annotating PubMed abstracts with concept occurrences. We chose the disease annotation task because a large, expert-crafted gold standard with information about inter-annotator agreement was available for comparison [10] and because we expected that the concept of "disease" would be a tractable place to start testing the ability of "the crowd" to process complex biomedical text.

### 2.1. *Guidelines for creating the NCBI Disease corpus*

To produce the original gold standard disease corpus, annotators were instructed to highlight a span of text, to identify a concept from the Unified Medical Language System (UMLS) that matched the meaning of the highlighted text and to assign it to one of four categories. The categories included: Specific Disease (e.g. "Diastrophic dysplasia"), Disease Class (e.g. "autosomal recessive disease"), Composite Mention (e.g. "Duchenne and Becker muscular dystrophy"), and Modifier ("e.g. colorectal cancer families"). Annotators completed this task using the PubTator interface [11]. They followed an annotation guideline document that contained a number of additional rules such as "annotate duplicate mentions" and "do not annotate overlapping mentions" [10]. The PubTator tool also automatically suggested annotations and provided direct access to a UMLS search tool.

It is worth noting here that these guidelines called for the annotation of a number of semantic types that are related to diseases but are not diseases in themselves. For example, the instructions said to highlight mentions of semantic types such as "Sign or Symptom" (e.g. "Back Pain") and "Acquired Abnormality" (e.g. "Hernia"). They also contained certain criteria that left room for annotator interpretation (and disagreement) such as the condition that the highlighted text contain "information that would be helpful to physicians and health care professionals". Further, the annotation rules occasionally depended on access to the UMLS system to make consistent judgments. For example the decision about whether to annotate the 'early onset' part of phrases like "early onset colorectal cancer" was likely determined by whether or not a relevant concept in the UMLS existed that included the "early onset" qualifier. These challenging aspects of the annotation guidelines, as well as their length (nearly 1000 words including the examples) and complexity, compelled us to write a new instruction set specifically designed for the AMT workers and the interface provided to them. Other than the wording and examples provided, the largest difference was that our instructions did not call for annotators to identify the type of mention or to identify the concept referred to by the mention in the UMLS.

## 2.2. Instructions and task description provided to AMT workers

The task posted on the AMT platform was described as: *"You will be presented with text from the biomedical literature which we believe may help resolve some important medically related questions. The task is to highlight words and phrases in that text which are diseases, disease groups, or symptoms of diseases. This work will help advance research in cancer and many other diseases!"*

After agreeing to perform the task, they were presented with the following instructions and asked to take a qualification test. Each rule in the instructions was followed by an animated GIF that illustrated how the correct annotations would look as they were created with our custom highlighting tool. In the tool, users click to highlight single words, click and drag to highlight spans of text, and click again to unhighlight a span. *"Here are some examples of correctly highlighted text. Please study these before attempting to take the qualification test. Please also feel free to refer back to these examples if you are uncertain. Instructions:"*

**1 Highlight all diseases and disease abbreviations**
"...are associated with Huntington disease ( HD )... HD patients received..."
"The Wiskott-Aldrich syndrome ( WAS ) , an X-linked immunodeficiency…"

**2 Highlight the longest span of text specific to a disease**
"... contains the insulin-dependent diabetes mellitus locus …"
and not just 'diabetes'.
"...was initially detected in four of 33 colorectal cancer families…"
and not just 'cancer'.
"...In inherited breast cancer cases…"
and not just 'breast cancer'

**3 Highlight disease conjunctions as single, long spans.**
"...the life expectancy of Duchenne and Becker muscular dystrophy patients.."
"... a significant fraction of breast and ovarian cancer , but undergoes…"

**4 Highlight symptoms - physical results of having a disease**
"XFE progeroid syndrome can cause dwarfism, cachexia, and microcephaly. Patients often display learning disabilities, hearing loss, and visual impairment."

**5 Highlight all occurrences of disease terms**
"Women who carry a mutation in the BRCA1 gene have an 80 % risk of breast cancer by the age of 70. Individuals who have rare alleles of the VNTR also have an increased risk of breast cancer ( 2-4 )".

**6 Highlight all diseases, disease groups and key disease symptoms**
"The set of 32 families in which no BRCA1 alterations were detected included 1 breast-ovarian cancer kindred manifesting clear linkage to the BRCA1 region and loss of the wild-type chromosome in associated tumors . Other tumor types found in BRCA1 mutation / haplotype carriers included prostatic, pancreas, skin, and lung cancer, a malignant melanoma, an oligodendroglioma, and a carcinosarcoma"

**7 Do not highlight gene names**

"... the spastic paraplegia gene (SPG) was found to.."

highlight only the disease mention, not the gene

"...Huntington disease (HD) is caused by variations in huntingtin (HTT)..."

the disease is highlighted, but the related gene is not.

"...Niethold-Alfred syndrome (NAS) is caused by mutation in the gene NAS…" .

In some cases the name of the gene may be the same as the name of the disease. In these cases you should only highlight the text if it is referring to the disease and not to the gene. The first NAS is highlighted while the second is not.

## 2.3. Worker task flow: Qualification test, 4 training examples, then real tasks

Before gaining access to the HITs, the workers had to pass a qualification test. This test was a series of true/false questions that assessed their comprehension of the annotation rules based on real examples. We allowed workers to pass the test with a fairly lenient threshold of 80% correct.

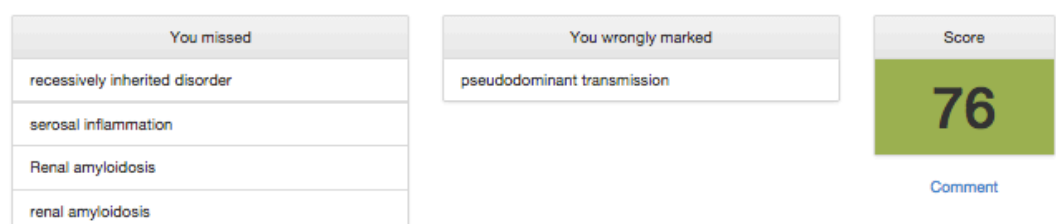

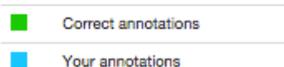

Figure 1: Feedback interface for annotators.

Once a worker earned the qualification, their first HIT was a demographic survey containing questions about: gender, age, occupation, education, and motivation. Following the survey, each worker progressed through the same four manually selected abstracts. After submitting their annotations for each of these four training documents, the system displayed their agreement with the gold standard annotations for that document in terms of an F measure. Figure 1 shows an example of the feedback screen that a worker would see after annotating a gold standard document. Workers were shown the annotations that they missed (false negatives), any that they incorrectly marked (false positives), and their F score for that document. Based on earlier studies we found that this feedback was an effective way to train workers to perform the task correctly, measurably improving performance with respect to simply providing the instructions. After the first four training documents were completed, the workers were presented with randomly selected abstracts to annotate. For 10% of the abstracts they annotated, feedback based on the gold standard was supplied to them. If the workers scored below in F measure of 0.5 on three gold standard documents in a row, they were blocked from continuing. When the workers annotated a non-gold standard document, they were presented with the annotations other workers (if any were present) on the same document for comparison.

### 2.4. *Data*

The experiment described here used the 593 abstracts in the Training Set of the NCBI Disease Corpus [10]. These were divided into three groups: the four training documents which all workers completed first and for which feedback was provided based on agreement with the gold standard annotations, an additional 10% (60) that were used to provide interspersed gold standard feedback, and the remaining 529 for which no gold standard-based feedback was provided. For documents not assigned to the gold standard group, workers were given feedback about the agreement between their annotations and those provided by other workers (if any were available). The results for each document are collected prior to the worker seeing any feedback and they only see each document once so, for the purposes of overall annotation quality assessment, all documents are treated quality.

### 2.5. *Quality statistics*

We compared the annotations generated by the AMT with the annotations in the NCBI Disease Corpus using strict matching. An annotation counted as a true positive only if it exactly matched the span of the corresponding gold standard annotation. For example, if the gold standard annotation was "early onset breast cancer" and the AMT annotation missed "early onset" but submitted "breast cancer", no partial credit for the overlap was given, and "breast cancer" would be considered a false positive. With exact matches as our comparison metric, we calculated true positives, false positives, and false negatives across all annotations in the dataset and used these to calculate global Precision (TP/(TP+FP)), Recall (TP/(TP+FN)) and F measure (2*P*R)/*P+R). (This method, referred to as micro-averaging, essentially treats the corpus as one large document.)

## 3. Results

### 3.1. *Workers*

Out of 346 that took it the qualification test, 145 workers passed and completed the annotation of at least one abstract. Work was distributed unevenly across the population, with 23 workers completing over 100 abstracts each, 80 completing over ten, and 65 completing 10 or less. Figure 2 shows the number of abstracts completed per worker. Only one worker was blocked from continuing after repeatedly performing poorly on the embedded gold standard test questions. That worker annotated 24 post-training abstracts with an average F-measure of 0.62 before being blocked.

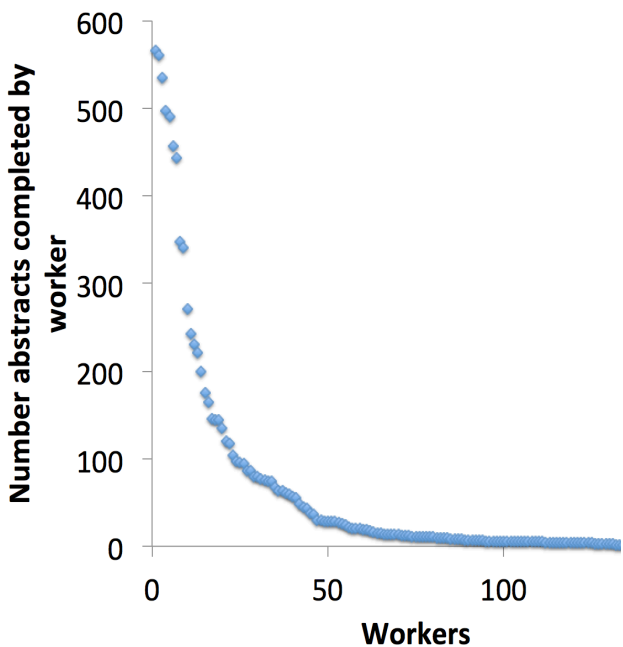

Figure 2: Number of abstracts processed per worker

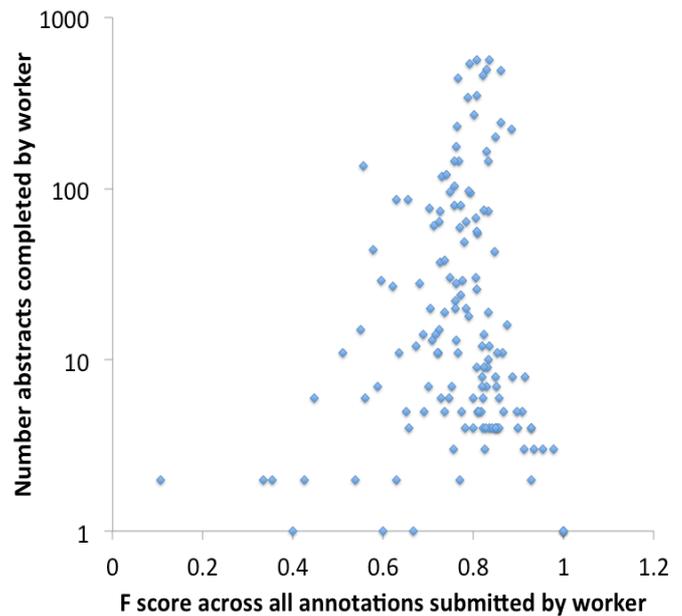

Figure 3: Quality of annotations for each worker compared to number of abstracts processed.

For this experiment, the average F score per worker across all of their annotations was 0.764 with a standard deviation of 0.130. Limiting to workers that processed more than ten abstracts the average F score increased only slightly to 0.767 with the standard deviation decreasing to 0.078. The average F score of workers that completed a hundred or more abstracts was 0.791 with a standard deviation of 0.078. This result supports the findings in Burger 2012, who observed a clear increase in performance for workers that completed more than 100 of their gene-mutation validation tasks. Figure 3 illustrates that overall, the workers generally performed at a high level for this task and that there was a weak positive correlation between the number of tasks performed and the quality and consistency of the work.

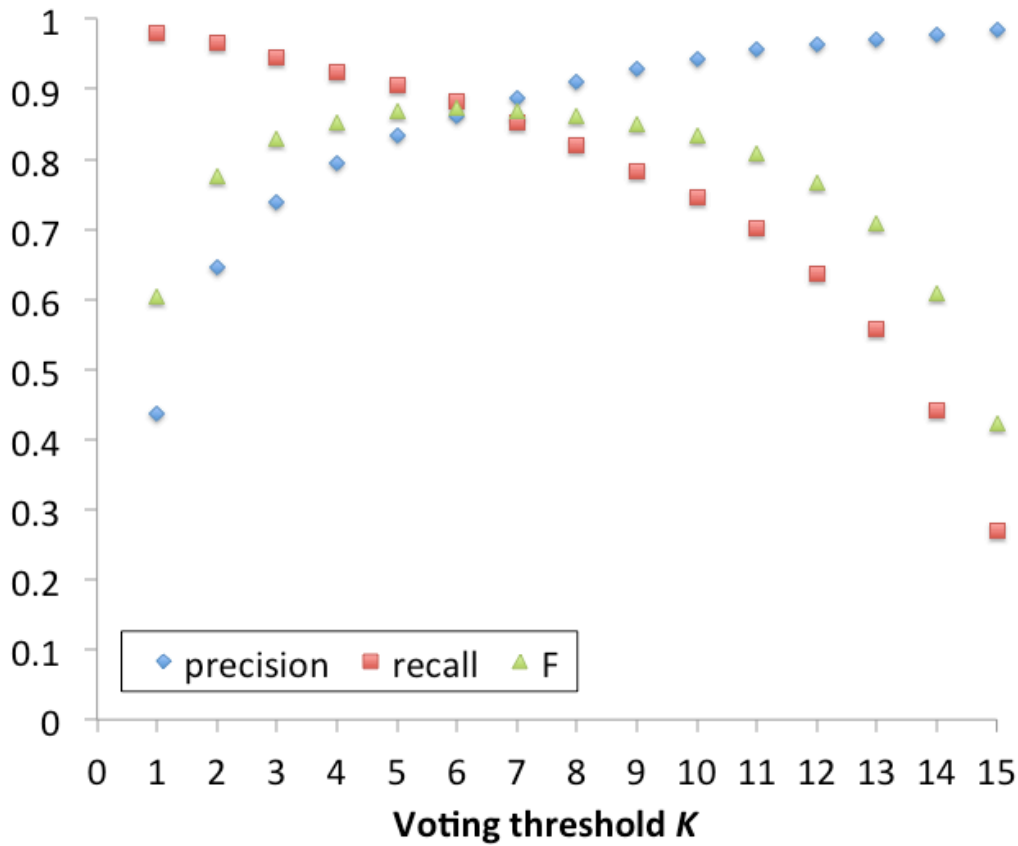

Figure 4: Impact of voting threshold $K$ on the level of agreement with the gold standard.

### 3.2. Wisdom of the crowd aggregation function

For this experiment, we collected annotations from 15 workers for each document (excluding the first four training examples which all workers saw). Following [8] and others, we employed a simple voting strategy to merge the annotations from multiple workers. For each annotation, we counted the number of different workers to produce that annotation and set a threshold K above which the annotation would be kept. We measured the quality of the annotation set generated based on all possible values of K as compared to the original gold standard documents. Figure 4 shows the Precision, Recall and F at each possible value of K. Precision varies from 0.436 at K=1 to 0.984 at K=15, Recall varies from 0.980 at K=1 to 0.269 at K=15, and the F measure peaks at 0.872 at K=6 (Precision = 0.862, Recall = 0.883).

### 3.3. Cost

At $.06 per HIT and 15 workers per document, we paid 90 cents per abstract. In addition, each worker was paid $.06 to take the demographic survey and $.24 to complete the first four training documents. 145 workers passed the qualification test and took part in this experiment. The total cost was thus 145*.30 (worker training) + 589*.90 (annotation) = $573.60. All 589 documents were annotated in a span of 9 days.

To gauge the impact of annotator redundancy (which equates to cost), we estimated the expected performance of the system with different numbers of workers (referred to as N) per document. (N for the complete collection, whose quality is depicted in Figure 4 was 15.) Using this data, we estimated performance at different values of N by sampling from the workers who completed each document. For example, to estimate performance at N = 6, for each document, we randomly selected the annotations from 6 of the workers who annotated that document, calculated the best value for the voting threshold K and recorded the associated F measure. For each value of N, we repeated this random process ten times and recorded the mean and standard deviation for the maximum F measure per N. As Figure 5 illustrates, the range of F scores varies from 0.78 at N = 1 to 0.87 at all N greater than 7. The largest increase in performance is between N = 2, average F = 0.78 and N = 3, where average F = 0.84.

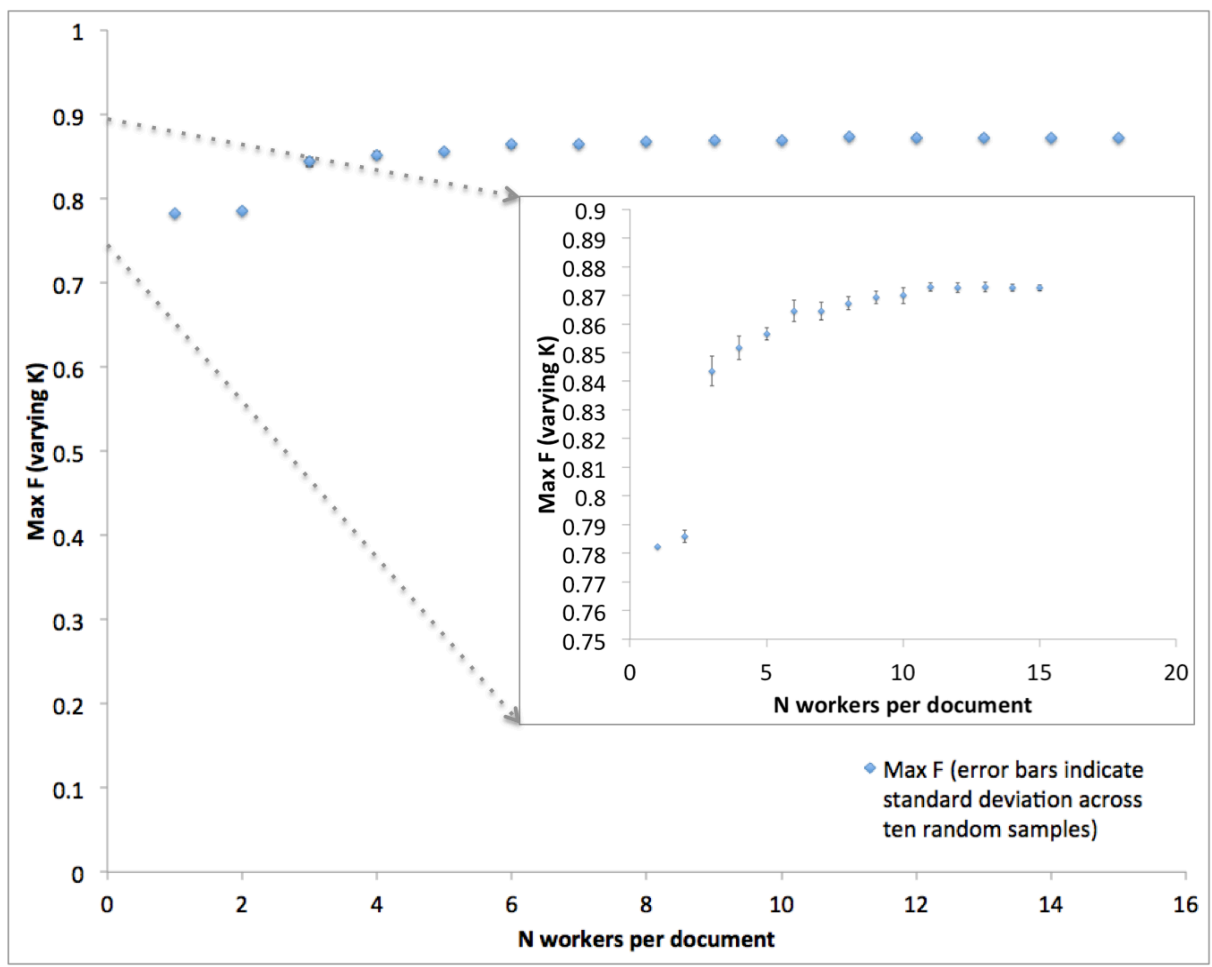

Figure 5: Impact of increasing the number of workers per abstract on annotation quality

### 3.4. *Machine learning*

We compared the performance of the BANNER machine learning system [12] given two training sets: a) the original gold standard training set (593 abstracts), and b) the annotations of the 589 abstracts considered here at the voting threshold K=6. Training the model with the original gold standard produced an F measure of 0.785 on the 100 abstract test set from the NCBI corpus (Precision 0.808 , Recall 0.764). Training the same system on the AMT-generated data produced an F measure of 0.739 (Precision 0.778 , Recall 0.703). While this model performed slightly worse on the selected test set than the gold-standard-trained system, it performed substantially better than prior approaches including a BANNER model trained on the original corpus AZDC Disease corpus (F = 0.35) and the NCBO Annotator using just the Human Disease Ontology (F = 0.26).

### 3.5. *Demographics*

According to the pre-task survey, the workers that contributed to this study were 59% female, 41% male. They had a mean age of 32 with 68% between 21 and 35, 19% between 36 and 45, and 11% 46 or older. Regarding education, workers were widely distributed including one worker that reported not finishing high school and four that completed PhD programs. The largest group (28%) had completed a four year college degree, with an additional 12% completing a masters program. 20% of our AMT workforce reported being unemployed, 15% were students, with the remaining 65% employed in a wide range of occupations with the top categories being "Technical" and "Science". In general, all of these trends are in agreement with prior demographic studies of the AMT worker population as a whole [13]. Asked why they worked on our HITs in particular, 85% selected "I want to make money", 69% selected "I want to help science" and 17% selected "Entertainment" (they could select multiple categories). Along with the consistent motivation to earn money, many workers expressed happiness in performing a task that (a) contributed to science and (b) involved them with subject matter that they found educational. This was reiterated in many emails received from the workers.

### 4. Discussion

One of the key rating-limiting factors in the advancement of the field of NLP is the production of annotated corpora. This is perceived to be even more the case for biomedical applications where the language is complex and full of jargon. Based on the results presented here, crowdsourcing appears to be a viable tool to help break down this bottleneck. This experiment demonstrated how a disease mention annotation corpus with more than 500 abstracts could be assembled for less than $600 in less than two weeks. Further, we provided evidence that the same system could produce a corpus with a similar level of quality in much less time and at much lower cost by reducing the number of redundant annotators per document (Figure 5). More sophisticated aggregation functions than voting (e.g. machine learning models) could likely further increase quality and reduce costs.

Crowdsourcing-based annotation systems should make it possible to create far more and far larger training sets than would be conceivable with the expert-only approach. But they will be different. Like the dataset generated here, most crowdsourced corpora will not reproduce gold standards exactly. In one sense this is a weakness. Existing NLP systems assume that training and testing data are perfect and binary. A span of text is a disease mention or it is not and there is no grey area. Evaluations of crowd-generated data, such as our assessment of the performance of the crowd-trained BANNER model, may thus show less favorable outcomes. However, in many cases in language the binary premise is likely not reflective of reality. There are always edge cases and ambiguities. Crowdsourced data offer the potential to identify the ambiguities of language and of annotation tasks in a computationally accessible way that could lead to important advances. To illustrate, consider the sentence : "*Significant differences were found between PWS patients, SIB controls, and WC controls in the prevalence of febrile convulsions, fever-associated symptoms, and temperature less than 94 degrees F*". In the gold standard, the only annotation captured is "PWS", an acronym for Prader-Willi syndrome. In the AMT results, 15 of 15 workers highlighted "PWS", but 13 of 15 workers also highlighted "febrile convulsions" and 11 highlighted "fever-associated symptoms". Both of these were considered false positives though in other documents highly similar terms such as "benign familial infantile convulsions" and "fever" were part of the gold standard annotations. Another example is provided in the sentence: "*We report the identification of a female patient with the X-linked recessive Lesch-Nyhan syndrome (hypoxanthine phosphoribosyltransferase [HPRT] deficiency)*". In the gold standard, the phrase "Lesch-Nyhan syndrome" is annotated. However, 11 of 15 AMT workers selected "X-linked recessive Lesch-Nyhan syndrome" instead. (In other gold standard annotations, "X-linked" and "recessive" are often included as modifiers in multi-word disease phrases.) Finally, consider the following title: "*von Willebrand disease type B: a missense mutation selectively abolishes ristocetin-induced von Willebrand factor binding to platelet glycoprotein Ib*". In the gold standard, both "von Willebrand disease type B" and the second "von Willebrand" are annotated. 13 of 15 workers highlighted the first occurrence correctly yet only 2 highlighted the second, resulting in a false negative. The next sentence of this abstract begins with "*von Willebrand factor (vWF) is a multimeric glycoprotein*" making it questionable whether the second "von Willebrand" was referring to the gene or the disease.

We do not bring these examples up to criticize the NCBI Disease Corpus itself. Arguments could be made for these decisions. The important point is that such edge cases occur in all corpus creation tasks. As a product of the multi-annotator perspective provided by (and in fact required by) crowdsourcing, ambiguity can be quantified directly. Emerging research is showing that such levels of disagreement are signals that can be used both during the training of machine learning models and in more realistic assessments of their predictions [14].

One of the key contributions of the work presented here is to solidify the potential of the crowd of people who will respond to an open call for participation to effectively process biomedical text. Here, we focused on the AMT worker population but other crowds might be tapped for future work. While the AMT workers were shown to be effective, the per-unit cost of this framework does puts constraints on the total amount of work that can be performed. Volunteer-based "citizen science" opens up the potential for far more work at lower cost [15]. As long as the collective mind of the crowd continues to outperform computational methods, it may even make sense to tap the crowd to directly perform information extraction tasks in support of biomedical research objectives rather than solely applying them to train computational systems.

**Acknowledgments**

We would like to thank all of the AMT workers. This research was supported by the National Institute of General Medical Sciences of the National Institutes of Health under award numbers R01GM089820 and R01GM083924, and by the National Center for Advancing Translational Sciences of the National Institute of Health under award number UL1TR001114.